\documentclass[conference]{IEEEtran}
\IEEEoverridecommandlockouts
\usepackage{cite}
\usepackage{amsmath,amssymb,amsfonts}
\usepackage{algorithmic}
\usepackage{graphicx}
\usepackage{textcomp}
\usepackage{xcolor}
\usepackage{amssymb}
\usepackage{ulem}
\usepackage{multicol}
\def\BibTeX{{\rm B\kern-.05em{\sc i\kern-.025em b}\kern-.08em
    T\kern-.1667em\lower.7ex\hbox{E}\kern-.125emX}}

\usepackage{bbm}
\newcommand{\Bmath}[1]{\mbox{\bf {#1}}}

\newcommand{\Bc}{\Bmath{c}}

\newcommand{\Bx}{\Bmath{x}}

\newcommand{\By}{\Bmath{y}}

\newcommand{\BW}{\Bmath{W}}

\newcommand{\BV}{\Bmath{V}}

\newcommand{\A}{{\cal A}}

\def\b{{\Bmath b}}

\def\s{{\Bmath s}}
\def\d{{\Bmath d}}

\def\H{{\Bmath H}}
\def\l{{\Bmath L}}

\def\A{{\Bmath A}}

\def\C{{\Bmath C}}

\def\V{{\Bmath V}}

\newtheorem{theorem}{Theorem}[section]

\newtheorem{definition}[theorem]{Definition}

\newtheorem{remark}[theorem]{Remark}

\numberwithin{equation}{section}
\newcommand{\norm}[1]{\left\lVert#1\right\rVert}

\usepackage{soul}

\author{
    \IEEEauthorblockN{Ziqiang Li\(^1\), Robert Simon Fong\(^2\), Kantaro Fujiwara\(^3\), Kazuyuki Aihara\(^1\), and Gouhei Tanaka\(^4\)}
    \IEEEauthorblockA{\(^1\)International Research Center for Neurointelligence, The University of Tokyo, Tokyo, 113-0033, Japan\\
    Email: \{ziqiang-li, kaihara\}@g.ecc.u-tokyo.ac.jp\\
                      \(^2\)School of Computer Science, University of Birmingham, Birmingham, B15 2TT, UK\\
                      Email: \{r.s.fong\}@bham.ac.uk\\
                      \(^3\)Graduate School of Medicine, The University of Tokyo, Tokyo, 113-0033, Japan\\
                      Email: \{kantaro\}@sat.t.u-tokyo.ac.jp\\
                      \(^4\)Department of Computer Science, Graduate School of Engineering, Nagoya Institute of Technology,\\ Nagoya, 466-8555, Japan\\
                      Email: \{gtanaka\}@nitech.ac.jp
                      }
}

\begin{document}
\title{Structuring Multiple Simple Cycle Reservoirs with Particle Swarm Optimization}
\maketitle

\begin{abstract}
Reservoir Computing (RC) is a time-efficient computational paradigm derived from Recurrent Neural Networks (RNNs). The Simple Cycle Reservoir (SCR) is an RC model that stands out for its minimalistic design, offering extremely low construction complexity and proven capability of universally approximating time-invariant causal fading memory filters, even in the linear dynamics regime. This paper introduces Multiple Simple Cycle Reservoirs (MSCRs), a multi-reservoir framework that extends Echo State Networks (ESNs) by replacing a single large reservoir with multiple interconnected SCRs. We demonstrate that optimizing MSCR using Particle Swarm Optimization (PSO) outperforms existing multi-reservoir models, achieving competitive predictive performance with a lower-dimensional state space. By modeling interconnections as a weighted Directed Acyclic Graph (DAG), our approach enables flexible, task-specific network topology adaptation. Numerical simulations on three benchmark time-series prediction tasks confirm these advantages over rival algorithms. These findings highlight the potential of MSCR-PSO as a promising framework for optimizing multi-reservoir systems, providing a foundation for further advancements and applications of interconnected SCRs for developing efficient AI devices.

\end{abstract}


\section{Introduction}

Recurrent Neural Networks (RNNs) are input-driven parametric state-space machine learning models designed to capture temporal dependencies in sequential input data streams. Time series data are sequentially encoded into this state space, allowing dynamic capture of temporal information via state-space vectors. 

Reservoir Computing (RC) is a subclass of RNNs where the state-space representation (the reservoir-encoder) is fixed and non-trainable, while only a static readout layer is trained. This design simplifies the training process by concentrating adjustments to the readout layer, avoiding backpropagation through time and improving computational efficiency. The simplest implementation of RC models includes Echo State Networks (ESNs) \cite{Jaeger2001, Maass2002, Tino2001}.

\par
ESNs have been applied successfully to various tasks \cite{Jaeger2004, Bush2005, Tong2007}. However, designing task-specific reservoir-encoders remains a practical challenge. This process often relies on trial-and-error \cite{Xue2007}, with limited strategies for selecting optimal reservoirs. Random connectivity and weights rarely yield optimal performance, and spectral radius constraints alone are insufficient for robust parameter tuning \cite{Ozturk2007}. Additionally, the high dimensionality of the coupling matrices complicates direct optimization for complex temporal data processing.
\par

\par
Simple Cycle Reservoirs (SCR) emerge as a specialized form of ESN models, characterized by a single degree of freedom in reservoir construction, utilizing uniform ring connectivity and binary input weights with an aperiodic sign pattern. SCRs have been shown to be universal approximators of time-invariant dynamic filters with fading memory over $\mathbb{C}$ and $\mathbb{R}$ \cite{li2023simple, fong2024universality}, respectively. Recent findings reveal that at the edge of stability, the kernel representation of SCR replicates the Fourier decomposition, providing a natural link between reservoir-based signal processing and classical signal processing models \cite{fong2024linear}. These properties make SCR highly suitable for integration into photonic circuits, enabling high-performance, low-latency processing \cite{bauwens2022using, larger2017high, harkhoe2020demonstrating}.

Furthermore, a predictive model \cite{tino2024} leveraging the kernel view of linear ESNs \cite{tino2020dynamical} driven by SCR dynamics has achieved forecasting performance comparable to state-of-the-art transformer models on univariate time series tasks. This highlights the practical capabilities of SCRs in advanced forecasting applications.

Recently, Multiple Reservoir Echo State Networks (MRESNs) have been proposed to address the challenge of optimizing high-dimensional connectivity matrices while also enhancing the computational capabilities of ESNs~\cite{gallicchio2017deep, li2022multi, li2023multi}. Instead of using a single large reservoir-encoder, MRESNs employ multiple interconnected smaller reservoir-encoders (vertex-encoders). This approach shifts the focus from optimizing large coupling matrices to designing the network topology governing the interconnections among the smaller vertex-encoder. Specifically, these interconnections, along with their coupling to input signals, can be represented as a Directed Acyclic Graph (DAG)~\cite{li2024designing}.

While studies~\cite{li2024designing,wcislo2021grouped} have empirically shown that optimizing connectivity via a DAG can improve the performance of MRESNs in certain time series processing tasks, this optimization has not been performed alongside other hyperparameters, such as input scaling across multiple reservoir encoders. As a result, existing multi-reservoir systems are likely suboptimal.

Furthermore, most studies \cite{li2021multi,li2022multi,li2023multi} focus exclusively on MRESNs with randomized reservoir encoders. These approaches cannot be directly applied to some physical multi-reservoir systems, as those physical devices rely on simple cycle reservoirs as encoders.

In this paper, we address both of these limitations. We adopt SCRs as the foundation of our multi-reservoir system, modeling both input scaling  and interconnections among multiple SCRs as a weighted directed acyclic graph (DAG). In particular, the interconnections between reservoirs, represented by the adjacency matrix of a graph, are optimized using Particle Swarm Optimization (PSO).

Through numerical simulations on three benchmark time-series prediction tasks, we demonstrated that our proposed multi-reservoir system with optimal structures and hyperparameters obtained through the weighted DAG optimization achieves competitive predictive performance compared to a single SCR and other multi-reservoir models. Moreover, this result is achieved with a lower-dimensional state space.

The main contribution of this paper is summarized as follows:
\begin{itemize}
    \item We propose \textbf{Multiple Simple Cycle Reservoirs (Multi-SCR or MSCR)} -- a novel multi-reservoir system that adopts SCRs as core vertex-encoders, organized through weighted DAG optimization.
    
    \item We achieve simultaneous optimization of both input scaling and connectivity of MSCR by introducing PSO.
    
    \item Experimental results demonstrate the effectiveness of our proposed model compared to existing methods across three benchmark time-series processing tasks.
    
\end{itemize}

\par
The rest of this paper is organized as follows: Section~\ref{Sec: Notion} presents the formal construction of MSCR.  Section~\ref{Sec: PSO} summarizes PSO on Euclidean spaces. Section~\ref{sec:experiments} presents the details of the numerical experiments. Finally,  Section~\ref{sec:discussion} concludes the paper with a discussion of our results.

\section{Notion of MSCR}
\label{Sec: Notion}

In this section, we introduce the notion of reservoir systems, Simple Cycle Reservoirs (SCR), and Multi-SCR (MSCR), which are focused on throughout the paper. 
\begin{definition}\label{def.lrc} A \textbf{reservoir system} over $\mathbb{R}$ 
is the quadruplets $R:= (\BW,\BW_{in},f,h)$ where the \textbf{state coupling} $\BW \in \mathbb{M}_{n\times n}\left(\mathbb{R}\right)$ is an $n\times n$ matrix over $\mathbb{R}$, the \textbf{input-to-state coupling} $\BW_{in} \in \mathbb{M}_{n\times m}\left(\mathbb{R}\right)$ is an $n\times m$ matrix, $f:\mathbb{R}^n \rightarrow \mathbb{R}^n$ is a fixed activation function, and the state-to-output mapping (\textbf{readout}) $h:\mathbb{R}^n \to \mathbb{R}^d$ is a (trainable) continuous function. 

The corresponding dynamical system is given by:
\begin{equation} \label{eq.system}
   \begin{cases} 
    \Bx_t &= f\left( \BW \Bx_{t - 1} + \BW_{in} u_t \right)\\
    \By_t &= h(\Bx_t)
    \end{cases}
\end{equation}
where $\{u_t\}_{t\in\mathbb{Z}_-} \subset \mathbb{R}^m$, $\{\Bx_t\}_{t\in\mathbb{Z}_-} \subset \mathbb{R}^n$, and $\{\By_t\}_{t\in\mathbb{Z}_-} \subset \mathbb{R}^d$ are the external inputs, states, and outputs, respectively.
We abbreviate the dimensions of $R$ by $(n,m,d)$. A reservoir system is \textbf{linear} if the activation function $f$ is the identity function, where we shall abbreviate $R$ by $R = (\BW,\BW_{in},h)$ 

We make the following assumptions for the system:
\begin{enumerate}
    \item $\mathbf{W}$ is assumed to be strictly \textbf{contractive}. In other words, its operator norm $\norm{\mathbf{W}}<1$. The system \eqref{eq.system} thus satisfies the fading memory property (FMP) \cite{li2023simple}.
    \item We assume {that} the input stream $\{u_t\}_{t\in\mathbb{Z}_-}$ is \textbf{uniformly bounded}. In other words, there exists a constant $M$ such that $\norm{u_t}\leq M$ for all $t\in\mathbb{Z}_-$. 
\end{enumerate}
 The contractiveness of $\BW$ and the uniform boundedness of the input stream imply that the images 
 $\Bx \in \mathbb{R}^n$ of the inputs $\Bc \in (\mathbb{R}^m)^{\mathbb{Z}_-}$ under the linear reservoir system live in a compact space $X \subset{\mathbb{R}}^n$. With slight abuse of mathematical terminology, we call $X$ a \textbf{state space}.
\end{definition}

\begin{definition} Let $\C=[c_{ij}]$ be an $n\times n$ matrix. We say that $\C$ is a \textbf{permutation matrix} if there exists a permutation $\sigma$ in the symmetric group $S_n$ 
such that 
\[c_{ij}=\begin{cases} 1, &\text{ if }\sigma(i)=j, \\0, &\text{ if otherwise.}\end{cases}\]
We say that a permutation matrix $\C$ is a \textbf{full-cycle permutation}\footnote{Also called left circular shift or cyclic permutation in the literature.} if its corresponding permutation $\sigma\in S_n$ is a cycle permutation of length $n$.  Finally, a matrix $\BW=\rho \cdot \C$ is called a \textbf{contractive full-cycle permutation} if $\rho\in(0,1)$ and $\C$ is a full-cycle permutation. 
\end{definition}

SCR is a special class of reservoir system \cite{rodan2010minimum} with a very small number of degrees of freedom, yet it manages to retain the performance capabilities of more complex or (unnecessarily) randomized constructions. 

\begin{definition}
\label{def.rc}
    A reservoir system $R = \left(\BW,\BW_{in},f,h\right)$ with dimensions $(n,m,d)$ is called a \textbf{Simple Cycle Reservoir (SCR)}
    \footnote
     {We note that the assumption on the aperiodicity of the sign pattern in $V$ is not required for this study.}
    if 
    \begin{enumerate}
        \item $\BW$ is a contractive full-cycle permutation, and
        \item $\BW_{in} \in \mathbb{M}_{n \times m}\left(\left\{-1,1\right\}\right)$. 
    \end{enumerate}
\end{definition}

Recently, it was shown that even with such a drastically reduced design complexity, \emph{linear} SCR models are universal approximators of fading memory filters \cite{li2023simple, fong2024universality}.

In this paper, we adopt SCRs as the core of our multi-reservoir system due to their demonstrated effectiveness and practical advantages. In particular, the motivation for choosing SCRs is fourfold:
\begin{itemize}
    \item \textbf{Universality:} SCRs are demonstrated to be universal approximators for general reservoir systems \cite{li2023simple, fong2024universality}.
    \item \textbf{Reduced randomness:} SCRs use fixed, deterministic state space representations, minimizing randomness and enhancing reproducibility.
    \item \textbf{Practical performance:} Method based on a single SCR has been shown to be effective for both univariate and multivariate time series forecasting \cite{tino2024}.
    \item \textbf{Hardware compatibility:}  The simple structure of SCRs makes them ideal for hardware implementations \cite{Appeltant2011InformationPU, NTT_cyclic_RC, bienstman2017, Abe2024May}.  This study lays the groundwork for advancing SCR-based architectures toward scalable and efficient hardware deployment.
\end{itemize}

\begin{figure*}[htbp]
    \centering
    \includegraphics[scale=0.5]{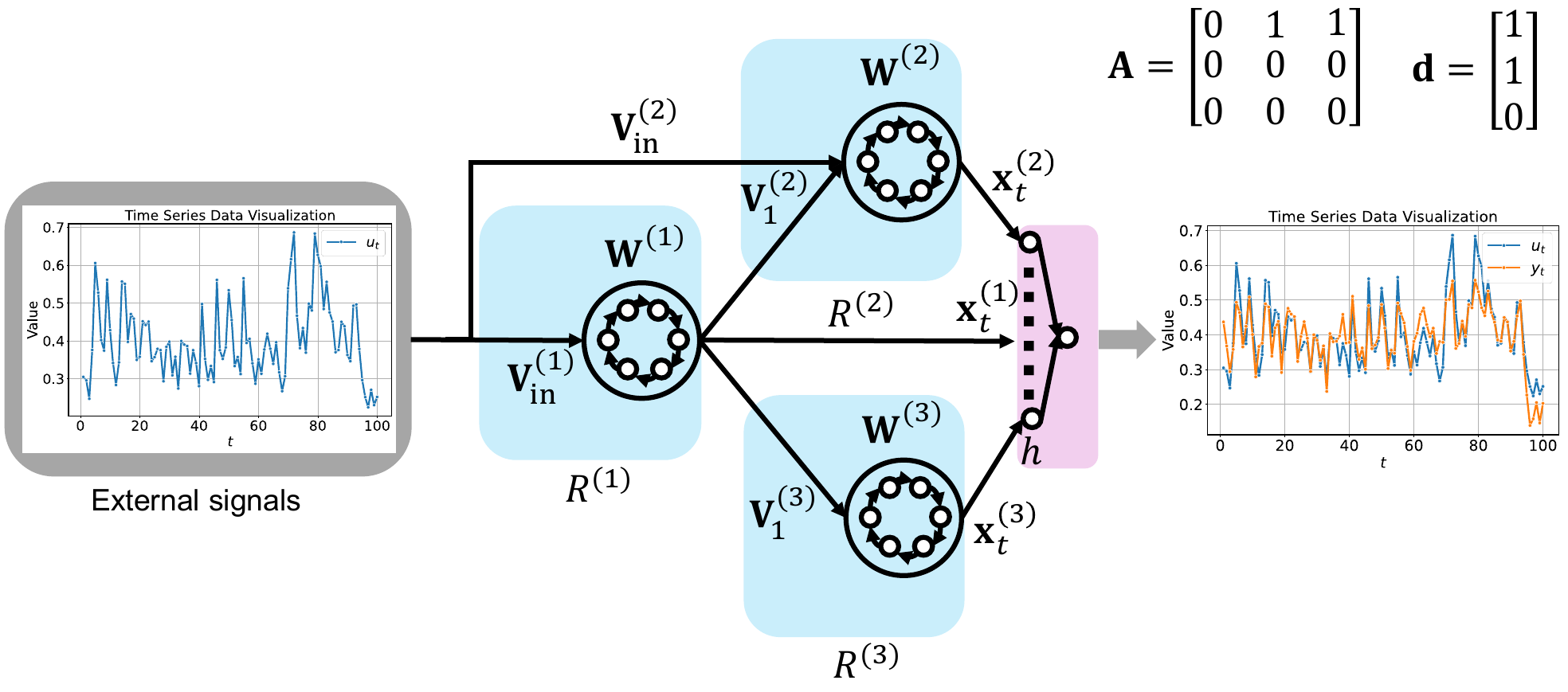}
    \caption{A schematic diagram of an MSCR of order three for dealing with time-series prediction tasks. In the shown case, all the vertex-encoder are valid, and the corresponding rank of the MSCR is three.}
    \label{fig:schematic_diagram}
\end{figure*}

We conclude the section by introducing the notion of multi-simple cycle reservoirs. 
\par
\begin{definition}
\label{def.MSCR}
Given positive integer $k > 1$, a {\bf Multi-Simple Cycle Reservoir of order $k$} (MSCR) is the system $S = \left(\{\BW^{(i)}\}_{i=1}^k, \{\mathcal{W}_{in}^{(i)}\}_{i=1}^k,\s, \H, \d, \A, f, h \right)$ with dimensions $(n, k, m, d)$. The system $S$ consists of $k$ interconnected \textbf{vertex-encoders}\footnote{Also called \textbf{encoders} in ~\cite{li2022multi,li2023multi}.} 
, denoted by $R^{(i)} = \left(\BW^{(i)}, \s_i \cdot \mathcal{W}_{in}^{(i)}, f, h \right)$, each with dimension $n$. The system $S$ is defined with the following conditions:
\begin{enumerate}   
    \item $\s \in \mathbb{R}^k$ denotes the set of input scaling factors for the external input, where each $\s_i$ corresponds to the scaling factor associated with $R^{(i)}$.
    
    \item $\H \in \mathbb{M}_{k\times k}(\mathbb{R})$ represents the matrix of input scaling factors between vertex-encoder. Specifically, $\H_{ij}$ denotes the scaling factor for input from $R^{(j)}$ to $R^{(i)}$.
    \item The connectivity between vertex-encoder is represented by a directed graph with node-adjacency matrix $\A = \left[\A_{ij}\right]_{i,j = 1}^k \in \mathbb{M}_{k \times k}(\mathbb{R})$, satisfying:
        \begin{itemize}
            \item $\A_{ii} = 0$ for all $i$, ensuring no self-loops.
            \item $\A_{ij} = 1$ if the output of the $i$-th vertex-encoder is used as input to the $j$-th vertex-encoder; otherwise, $\A_{ij} = 0$.
        \end{itemize}
    \item $\d \in \mathbb{M}_{k \times 1}\left(\left\{0,1\right\}\right)$ indicates whether each vertex-encoder receives the external input, where:
        \begin{itemize}
            \item $\d_i = 1$ if the $i$-th vertex-encoder receives the external input.
            \item $\d_i = 0$ otherwise.
        \end{itemize}

    \item Each vertex-encoder $R^{(i)} \in S$ is described as follows:
    \begin{itemize}
        \item The coupling matrix $\BW^{(i)} \in \mathbb{R}^{n \times n}$ is a contractive full-cycle permutation matrix, for $i = 1, 2, \dots, k$.
        \item The state of $R^{(i)}$ at time $t$ is denoted by $\Bx_t^{(i)}$. The input $\mathcal{U}_t^{(i)} \in \mathbb{R}^{m+n\cdot\left(k-1\right)}$ to $R^{(i)}$ at time $t$ is formed by concatenating the external input $u_t$ with the states of all \emph{other} vertex-encoders:
        \[
        \mathcal{U}_t^{(i)} = \left(u_t, \Bx_t^{(1)}, \dots, \Bx_t^{(i-1)}, \Bx_t^{(i+1)}, \dots, \Bx_t^{(k)}\right).
        \]
        \item $\mathcal{W}_{in}^{(i)}$ is a linear map that transforms $\mathcal{U}_t^{(i)}$ into the state space of $R^{(i)}$ by: 
        \begin{align*}        \mathcal{W}_{in}^{(i)}\left(\mathcal{U}_t^{(i)}\right) &= \left(\V_{in}^{(i)} \d_i \s_i u_t \right. \\
        & \quad \quad \left. + \sum_{\substack{j = 1 \\ j \neq i}}^k \BV_j^{(i)}  \H_{ji}\A_{ji} \Bx_t^{(j)} \right),
        \end{align*}
        where:
        \begin{itemize}
            \item $\V_{in}^{(i)} \in \mathbb{M}_{n \times m}\left(\{-1, 1\}\right)$ is the input-to-state coupling map of the external input.
            \item $\BV_j^{(i)} \in \mathbb{M}_{n \times n}\left(\{-1, 1\}\right)$ is the input-to-state coupling map from the state of $R^{(j)}$ to $R^{(i)}$.
        \end{itemize}
        \item Both $\V_{in}^{(i)}$ and $\BV_j^{(i)}$ are implicitly defined by $\mathcal{W}_{in}^{(i)}$.
    \end{itemize}

    \item The dynamics of the $i$-th vertex-encoder $R^{(i)}$, for $i = 1, 2, \dots, k$, are defined as:
    \begin{align*}
    \Bx^{(i)}_t &= f\left(\BW^{(i)} \Bx^{(i)}_{t-1} + \mathcal{W}_{in}^{(i)}\left(\mathcal{U}_t^{(i)}\right) \right) 
    \end{align*}
    where:
    \begin{itemize}
        \item $f$ is the activation function; all vertex-encoders share the same activation function.
        \item $\{\mathcal{U}^{(i)}_t\}_{t \in \mathbb{Z}_-}$ is the concatenated input sequence.
        \item $\{\Bx^{(i)}_t\}_{t \in \mathbb{Z}_-} \subset \mathbb{R}^n$ is the state sequence of the $i$-th vertex-encoder.
    \end{itemize}
    
    \item The concatenated state vector $\Bx \in \mathbb{R}^{k \cdot n}$ at time $t$ is formed by combining the states of all $k$ vertex-encoder: 
        \[
        \Bx_{t} = \left(\Bx^{(1)}_{t}, \dots, \Bx^{(k)}_{t}\right), \quad \text{where } \Bx^{(i)}_{t} \in \mathbb{R}^n.
        \]
    The system output sequence is given by $\{\By_t\}_{t \in \mathbb{Z}_-} \subset \mathbb{R}^d$. The readout function $h$ operates on the global state $\Bx_t$ or a linear combination of component states: 
    \[
    h(\Bx_t) = h\left(\sum_{i=1}^k a_i \cdot \Bx^{(i)}_t \right),
    \]
    where $a_i \in \mathbb{R}$ are mixing coefficients.
\end{enumerate}
\end{definition}

Consider a directed graph $G = (V, E)$ with $\vert V \vert = k$ and a vertex $i \in V$. Recall the \textbf{reachability matrix} $\mathcal{R} = \left[\mathcal{R}_{ij}\right]_{i,j = 1}^k$ is a $k \times k$ matrix over $\left\{0,1\right\}$ defined as:
\begin{itemize}
    \item $\mathcal{R}_{ij} = 1$ if there exists a path from node $i$ to node $j$, 
    \item $\mathcal{R}_{ij} = 0$ otherwise.
\end{itemize}

The reachability matrix $\mathcal{R}$ can be computed from the connectivity matrix $\A$ using the Floyd-Warshall algorithm \cite{warshall1962theorem} with $O(k^3)$. 



\begin{definition}
    Given an MSCR $S = \left(\{\BW^{(i)}\}_{i=1}^k, \{\mathcal{W}_{in}^{(i)}\}_{i=1}^k, \s, \H, \d, \A, f, h \right)$ of order $k$, let $\mathcal{R}$ denote the reachability matrix corresponding to the directed graph defined by $\A$. A vertex-encoder $R^{(i)} \in S$ is \textbf{valid} if there exists $j = 1,\ldots,k$ such that:
    \begin{align*}
    \mathcal{R}_{ji} \cdot D_j = 1.
    \end{align*}
    Otherwise, a vertex-encoder is \textbf{invalid}. 
    \par
    The \textbf{rank} of $S$ is the number of valid vertex-encoders of the system $S$, given by:
    \begin{align*}
        \operatorname{rank}(S) := 
        \sum_{i=1}^k \mathbbm{1}\left(
        \sum_{j=1}^k \mathcal{R}_{ji}\cdot \d_j
        \right),
    \end{align*}
    where $\mathbbm{1}(x)$ is the indicator function:
    \[
    \mathbbm{1}(x) =
    \begin{cases} 
    1 & \text{if } x \neq 0, \\
    0 & \text{if } x = 0.
    \end{cases}
    \]
\end{definition}

\begin{remark}
\label{rmk:rank}
By construction an MSCR of order $k$ with rank $1$ is equivalent to a single SCR for any $k \in \mathbb{N}_+$.
\end{remark}

For the rest of the paper, we will consider reservoir systems operating on univariate input, i.e., the input dimension is $m = 1$. The input and target are denoted by $\{u_t\}_{t\in\mathbb{Z}_-} \subset \mathbb{R}$ and $\{\By(t)\}_{t\in\mathbb{Z}_-} \subset \mathbb{R}$, respectively. The spectral radius of the $i^{th}$ vertex-encoder $R^{(i)}$ in a Multi-Simple Cycle Reservoir of order $k$ will be denoted by $\rho^{(i)}$. In this work, we consider the data flow in the MSCR follows the rules listed as follows:

A schematic diagram of MSCR is illustrated in Figure~\ref{fig:schematic_diagram}.


\section{Particle Swarm Optimization}
\label{Sec: PSO}
\textbf{Particle Swarm Optimization} (PSO) \cite{eberhart1995particle} on Euclidean space $\mathbb{R}^n$ is a population-based meta-heuristic optimization algorithm. PSO is a zeroth-order algorithm motivated by swarm intelligence. PSO makes minimal assumptions on the search space -- it does not require the objective function to be differentiable, and only pairwise addition and negation are required for the search space. 

In this section, we follow the exposition of \cite{borckmans2010modified, fong_population-based_2022} and describe the evolution dynamics of PSO on Euclidean space $\mathbb{R}^n$ \cite{eberhart1995particle}. PSO starts with $N$ randomly initialized search agents (particles) in $\mathbb{R}^n$. Inspired by swarm intelligence, each particle $x_i^\ell$ `evolves' in $\mathbb{R}^n$ according to the following equations in the subsequent $\ell^{\text{th}}$ iteration:

\begin{align}
    v_i^{\ell+1} &= \underbrace{w^\ell\cdot v_i^\ell}_{\text{inertia}} + \underbrace{c\cdot \alpha_i^\ell \left( y_i^\ell - x_i^\ell\right)}_{\text{nostalgia}} + \underbrace{s\cdot \beta_i^\ell \left(\hat{y}^\ell - x_i^\ell \right)}_{\text{social}} \label{eqn:pso:1}\\
    x_i^{\ell+1} &= x_i^\ell + v_i^{\ell+1} \label{eqn:pso:2} \quad .
\end{align}

The velocity of each particle in the PSO dynamics (Equation~\eqref{eqn:pso:1}) consists of three distinct components, each serving a specific purpose. The inertia component preserves the particle's search direction from the previous iteration, the nostalgia component encourages particles to move toward their personal best-known positions, and the social component drives particles toward the global best-known position in the swarm.

The symbols of Equations~\eqref{eqn:pso:1} and Equation~\eqref{eqn:pso:2} are summarized as follows. These parameters collectively define the PSO dynamics, where particles update their positions $x_i^k$ based on velocities $v_i^{\ell+1}$.

\begin{itemize}
    \item \textbf{Particle Position ($x_i^\ell$):} Position of the $i^{\text{th}}$ particle in the search space at iteration $\ell$.

    \item \textbf{Particle Velocity ($v_i^\ell$):} Search direction (overall velocity) of the $i^{\text{th}}$ particle at iteration $\ell$.

    \item \textbf{Swarm Size ($N$):} Total number of particles in the swarm. In particular, $N$ determines the level of parallel exploration and exploitation of the optimization process within the search space.

    \item \textbf{Inertial Coefficient ($w^\ell$):}  weight of the inertia component, a predefined real-valued function on the iteration counter $\ell$.

    \item \textbf{Nostalgia Weight ($c$):} Weight of the nostalgia component, a pre-defined real number.

    \item \textbf{Social Weight ($s$):} Weight of the social component, a pre-defined real number.

    \item \textbf{Random Numbers ($\alpha_i^\ell, \beta_i^\ell$):} Independent random variables over $[0,1]$ for each particle and iteration, adding stochasticity to the nostalgia and social components to prevent premature convergence \cite{eberhart1995particle}.

    \item \textbf{Personal Best ($y_i^\ell$):} Best position discovered by the $i^{\text{th}}$ particle up to iteration $\ell$.

    \item \textbf{Global Best ($\hat{y}^\ell$):} Best position found by any particle in the swarm up to iteration $\ell$.
\end{itemize}

\section{Experiments}
\label{sec:experiments}
We designed experiments to systematically perform  the following key comparisons:

\begin{enumerate}
    \item Compare performance improvement:
    \begin{enumerate}
        \item MSCR (with optimized network topology and input scaling) 
        \item SCR (with optimized input scaling)
    \end{enumerate}
    on benchmark time-series prediction tasks.

    \item Compare performances of:
    \begin{enumerate}
        \item MSCR with optimized network topology
        \item MSCR with fixed network topology
    \end{enumerate}
    on benchmark time-series prediction tasks.

    \item Compare the effectiveness of:
    \begin{enumerate}
        \item The proposed PSO-based network topology optimization method
        \item Existing topology-search methods for MRESN
    \end{enumerate}
    in improving performance on benchmark time-series prediction tasks.
\end{enumerate}

\subsection{Model selection}
In this work, we utilize the PSO method introduced in Section~\ref{Sec: PSO} to optimize the input scaling factors and network topology of an MSCR. Specifically, given an MSCR $S = \left(\{\BW^{(i)}\}_{i=1}^k, \{\mathcal{W}_{in}^{(i)}\}_{i=1}^k, \s, \H, \d, \A, f, h \right)$ of order $k$, we optimize $\s$, $\H$, $\d$, and $\A$.

We denote an MSCR optimized via PSO as MSCR-PSO. For the optimization of $\d$ and $\A$, the particle positions at the 
$l$-th iteration is converted to binary values by mapping all positive values as one and negative values as zero.

The following RC models are used for comparison with MSCR-PSO: Simple Cycle Reservoir (SCR), Grouped Echo State Network with SCR vertex-encoders (GroupedSCR)\cite{gallicchio2017deep}, Deep Echo State Network with SCR vertex-encoders (DeepSCR)\cite{gallicchio2017deep}, and Multiple Simple Cycle Reservoir with an optimized network topology searched by a Genetic Algorithm (MSCR-GA)~\cite{li2024designing}.

Note that the input scaling factors for SCR, GroupedSCR, and DeepSCR were optimized using PSO, whereas those for MSCR-GA were pre-determined. These models were selected for their relevance to the research questions and distinct characteristics, as explained below:

\begin{itemize}
    \item \textbf{SCR} represents a \emph{single} reservoir system with minimal construction complexity, which has shown remarkable predictive performance in \cite{tino2024}. Comparing the predictive performance of SCR to that of MSCRs demonstrates whether multiple interconnected reservoir vertex-encoders can enhance the system's computational capability.
    \item \textbf{GroupedSCR} and \textbf{DeepSCR} are classical MSCRs with fixed network topologies. These models serve as baselines for evaluating whether optimized network topologies improve the computational ability of an MSCR, specifically in comparison to MSCR-PSO. 
    \item \textbf{MSCR-GA} builds upon recent work using a GA-based method to optimize only the network topology of an MRESN with pre-determined input and inter-scaling factors. In this study, we replaced the randomly initialized vertex-encoders with the SCR vertex-encoder to construct MSCR-GA. This comparison aims to measure the performance gains achieved by simultaneously optimizing scaling and topology factors in MSCR-PSO.
\end{itemize}

Each algorithm optimizes different parts of the construction of MSCR, namely the external input scaling vector~$\s$, the state scaling matrix~$\H$, the input-to-reservoir adjacency vector~$\d$, and the reservoir-to-reservoir adjacency matrix~$\A$. Table~\ref{Tab: situation} summarizes the parameters optimized in each model.

\begin{table}[htbp]
\centering
\caption{Summary of optimized $\s$, $\H$, $\d$, and $\A$ for each tested model.
}
\label{Tab: situation}
\scalebox{0.9}{
\begin{tabular}{ccccc}
\hline
Model    & Optimized $\mathbf{s}$ & Optimized $\mathbf{H}$ & Optimized $\mathbf{d}$ & Optimized $\mathbf{A}$\\ \hline
\hline
SCR      & \checkmark               &                            \\
DeepSCR   & \checkmark               &    \checkmark     &    $\times$      &    $\times$                 \\
GroupedSCR   & \checkmark               &     \checkmark  &    $\times$      &    $\times$                    \\
MSCR-GA  &    $\times$                     & $\times$  &    \checkmark      &    \checkmark                \\
MSCR-PSO & \checkmark               & \checkmark    & \checkmark               & \checkmark              \\ \hline
\end{tabular}
}
\end{table}
\par
For the numerical experiments, we introduce a bias term in the dynamics of MSCR, such as in \cite{rodan2010minimum}. In particular, for each vertex-encoder of an MSCR, denoted by $R^{(i)}:= (\BW^{(i)},\s_i \cdot \mathcal{W}_{in}^{(i)},f^{(i)},h^{(i)})$, a small bias vector $\b^{(i)}$ is introduced to the state evolution. In particular, the dynamics of the reservoir system now reads:

\begin{align*} 
     \Bx^{(i)}_t &= f\left(\BW^{(i)} \Bx^{(i)}_{t-1} + \V_{in}^{(i)} \d_i\s_i\cdot  u_t \right. \\
     & \quad \quad  \left.+ \sum_{\substack{j = 1 \\ j \neq i}}^k \BV_j^{(i)} \A_{ji} \Bx^{(j)}_t + 10^{-5} \cdot \b^{(i)} \right) 
\end{align*}
where $\b^{(i)} \in \mathbb{M}_{n\times 1}\left(\left\{-1,1\right\}\right) $ is a boolean vector generated either by the $\left(n+1\right)^{th}$ to the $\left(2n\right)^{th}$ digit of binary expansion of $\pi$ or sampled randomly using Bernoulli trails.

\subsection{Parameter settings}
\label{subsec:parameter settings}
We fixed the dimension of the state space for an SCR at 1000. For an MSCR model, we set the maximal order $k^{max}$ to 10, and the mapping space of each vertex-encoder to $n = 100$. We set the spectral radius for each vertex-encoder to $\rho = 0.95$ and used ridge regression with a regularization factor 1E-4 for training the readout function $h(\cdot)$. Note that the hyperparameters we searched in the experiments are initialization methods of $\mathbf{V}_{in}^{i}$ and $\mathbf{V}^{j}_{i}$ for $i=1,...,k$, and types of activation function $f$.
Coupling matrices $\mathbf{V}_{in}^{i}$ and $\mathbf{V}^{j}_{i}$ are initialized by either the binary expansion of $\pi$ of the appropriate size or sampled randomly using Bernoulli trails over $\left\{-1, 1\right\}$. The activation function was selected from the identity function and the hyperbolic tangent function. We recorded the average prediction performances across ten trials when the Bernoulli-based initialization method was adopted.

\par
For the settings of PSO, both the maximum number of iterations $l_{max}$ and the number of search agents $N$ were set to $100$. The inertial coefficient at the $l^{\text{th}}$ iteration was defined as $w^{l} = 0.5 + 0.5 \times \left(1 - \frac{l}{l_{max}}\right)$. The nostalgia weight $c$ and social weight $s$ were kept constant at $c = s = 2$. At each $k^{\text{l}}$ iteration, random numbers $\alpha_{i}^{l}$ and $\beta_{i}^{l}$ for each search agent were sampled from a uniform distribution over $\left[0, 1\right]$. For the settings of MSCR-GA, we adopted the default configuration described in \cite{li2024designing}.
\begin{figure*}[htbp]
    \centering
    \includegraphics[scale=0.45]{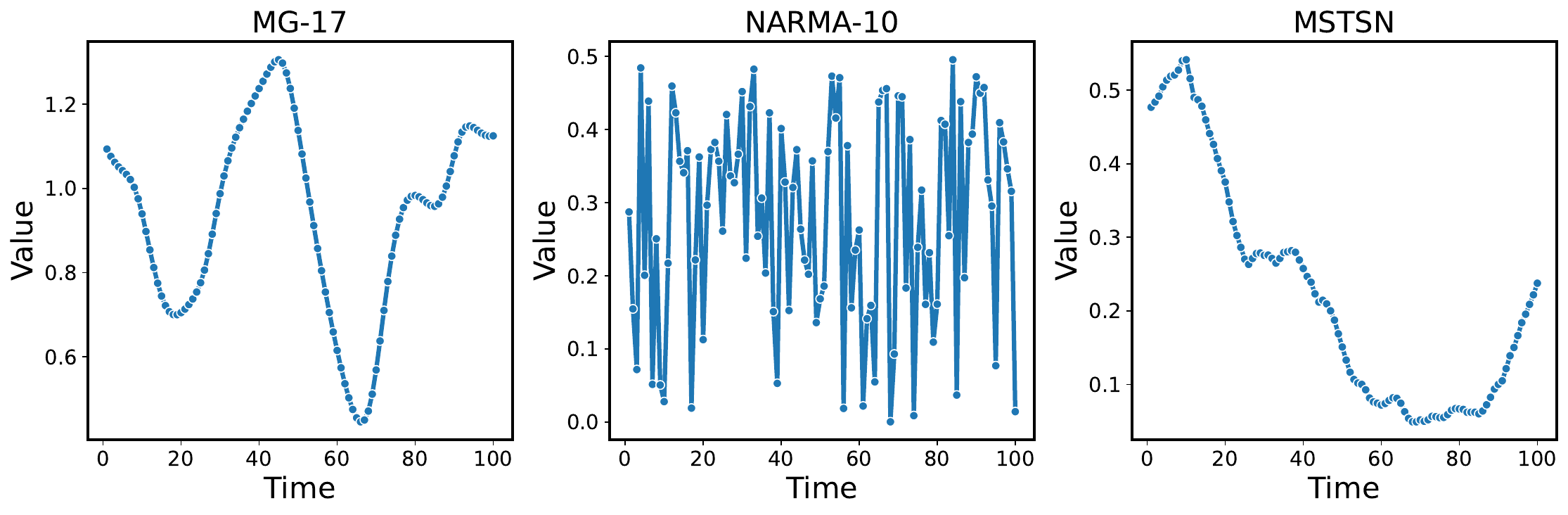}
    \caption{Glimpses of three time-series datasets: MG-17 (left), NARMA-10 (middle), MSTSN (right).}
    \label{fig:three_datasets}
\end{figure*}
\subsection{Datasets and task settings}
We evaluated the predictive capacity of the models using three benchmark datasets: the Mackey-Glass (MG) system, the tenth-order Nonlinear Autoregressive-Moving Average (NARMA-10) system, and the Monthly Smoothed Total Sunspot Number (MSTSN) dataset.

\par
The MG system with order $\tau$ is represented as follows:
\begin{equation}
    z_{t+1} = z_{t} + 0.1\left ( \frac{0.2z_{t-\tau/0.1}}{1+\left ( z_{t-\tau/0.1} \right )^{10} }-0.1z_{t}  \right) , 
\end{equation}
{where $z_{t}$ is the state variable at time $t$ and $\tau$ is the delay parameter.}
We set $\tau=17$ (MG-17)  to generate chaotic time series and executed an 84-step-ahead prediction task with $u_{t} = z_{t}$ and $y_{t} = z_{t+84}$~\cite{jaeger2004harnessing}. 
\par
The NARMA-10 system is described as follows: 
\begin{align}
z\left ( t+1 \right ) & = 0.3 z_{t} + 0.05 z_{t}\sum_{i = 0}^{n-9}z_{t-i}\\ \nonumber
&+ 1.5 \mu_{t-9} \mu_{t} + 0.1.
\end{align}
The external signal $\mu_{t}$ is randomly chosen from a uniform distribution $\left [ 0,0.5 \right ]$. The output $z_{t}$ is initialized by zeros for the first ten steps. We performed a one-step-ahead prediction task with $u_{t} = \mu_{t}$ and $y_{t} = z_{t+1}$.

\par
The MSTSN is a real-world univariate time-series dataset used for testing the prediction ability of a machine learning model~\cite{sidc}. It provides a 13-month smoothed average of monthly sunspot numbers recorded from January 1749, offering a clearer view of solar activity trends. We normalized all the time series into $\left[0, 1\right]$ as $u'\left( t \right)  =  \frac{u_t-\min \left ( u \right ) }{\max \left ( u \right ) - \min \left ( u \right )}$. We set a one-step-ahead prediction task to evaluate the computational ability of all the engaged models. 
\par
For all datasets, we partitioned the total data length into the lengths of the washout, training, validation, and test sets, which were set at $100$, $1000$, $1000$, and $1000$, respectively. The glimpses of datasets are shown in Fig.~\ref{fig:three_datasets}.

\subsection{Evaluation metric}
The Root Mean Square Error (RMSE) was used to evaluate the prediction performances of all the tested models, which can be formulated as follows:
\begin{equation}
    \mathrm{RMSE} = \sqrt{\frac{1}{N_{T}} \sum_{t=1}^{N_{T}}\left ( \hat{y}(t)- y(t)\right )^{2}}.
\end{equation}

\subsection{Results}
\label{subsec:results}
\begin{figure*}
    \centering
    \includegraphics[scale=0.8]{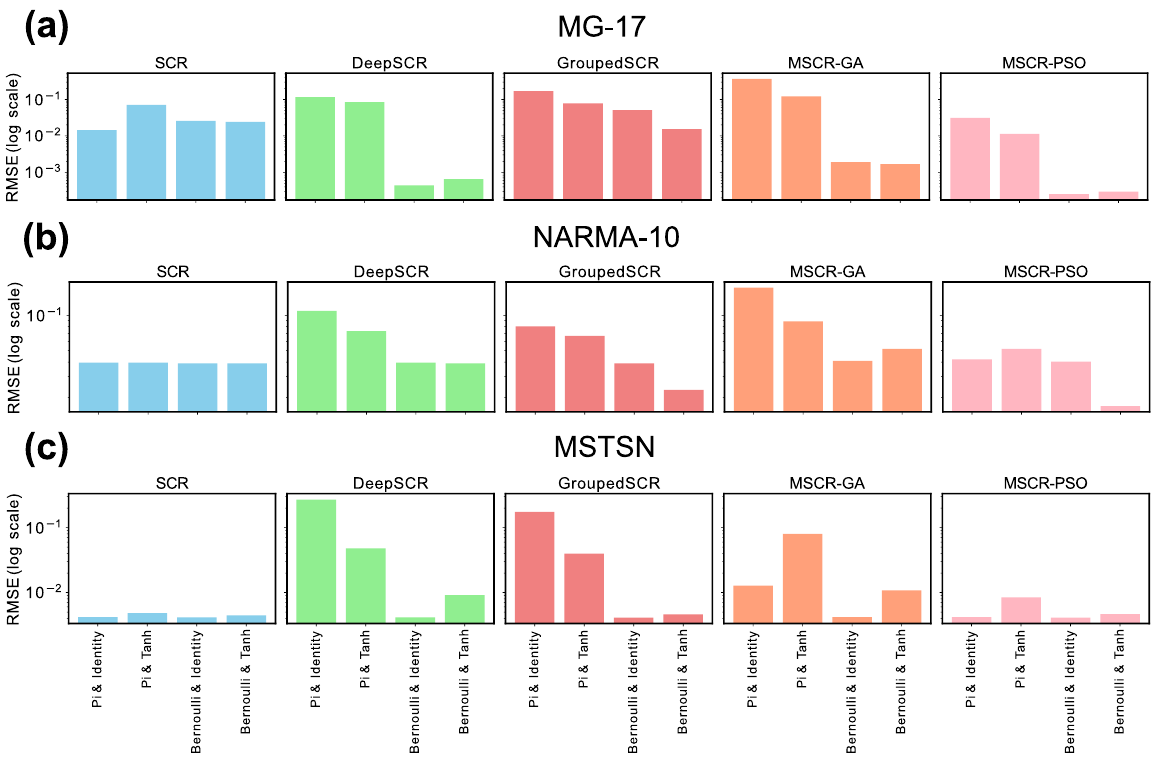}
    \caption{The best average prediction performances of all tested models on three datasets. (a) MG-17, (b) NARMA-10, and (c) MSTSN. }
    \label{fig:all_performances}
\end{figure*}
\begin{figure*}[htbp]
    \centering
    \includegraphics[scale=0.85]{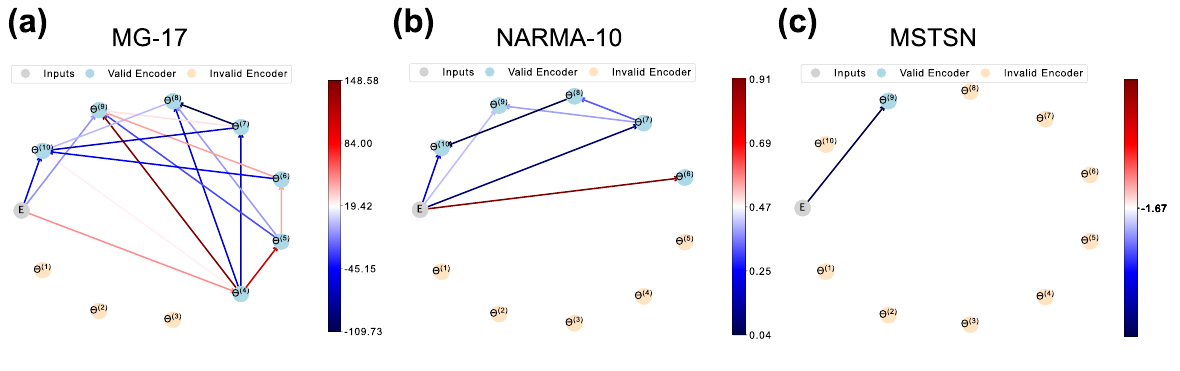}
    \caption{Visualization of searched network topologies (controlled by $\mathbf{d}$ and $\mathbf{A}$) and input scalings (controlled by $\mathbf{s}$ and $\mathbf{H}$) corresponding to the given prediction tasks: MG-17 (left), NARMA-10 (middle), and MSTSN (right). } 
    \label{fig:searched_topologies}
\end{figure*}

We conducted a systematic numerical analysis of all models (SCR, DeepSCR, GroupedSCR, MSCR-GA, and MSCR-PSO) across the parameters and configurations described in Section~\ref{subsec:parameter settings}. The average performance of all models across different hyperparameter settings is visualized in the bar graphs shown in {Figure~\ref{fig:all_performances}}.
The best performance and the corresponding choices of optimal hyperparameter settings (optimal choice of distribution and activation function) are summarized in Tables~\ref{tab: MG-17}-\ref{tab: MSTSN}. The network topologies of MSCR-PSO corresponding to its best performances are shown in Figure~\ref{fig:searched_topologies}.
\par
From Figure~\ref{fig:all_performances}(a), we observe that MSCR-PSO with a Bernoulli distribution and identity activation function achieves superior predictive performance on the MG-17 dataset. Similarly, as shown in Figure~\ref{fig:all_performances}(b), MSCR-PSO with a Bernoulli distribution and Tanh activation exhibits clear performance advantages on the NARMA-10 dataset. On the MSTSN dataset, the combination of a Bernoulli distribution and identity activation ensures relatively competitive computational performance across models, as illustrated in Figure~\ref{fig:all_performances}(c).

\par

Tables~\ref{tab: MG-17}-\ref{tab: MSTSN} present the models that achieved the best performance on the three datasets, along with their corresponding optimal hyperparameter settings, which include the choice of distribution and activation function. 

Overall, MSCR-PSO outperforms a single SCR across all three datasets, which indicates that an MSCR with several vertex-encoders organized by an optimized network topology is more effective than a single large SCR.
\par

From Tables~\ref{tab: MG-17}-\ref{tab: NARMA-10}, we observe that MSCR-PSO outperforms the second-best model on the MG-17 and NARMA-10 datasets in terms of RMSE by approximately 43.69\% and 26.84\%, respectively. Meanwhile, as shown in Table~\ref{tab: MSTSN}, MSCR-PSO achieves a prediction performance only 0.24\% below that of GroupedSCR. Note that for the MSTSN dataset, only a single vertex-encoder remains valid in MSCR-PSO, reducing its effective representation space dimension to just 10\% of that of GroupedSCR (see Figure~\ref{fig:all_performances}(c)).

Furthermore, MSCR-PSO demonstrates substantial improvements of approximately 85.21\%, 58.66\%, and 1.67\% compared to MSCR-GA on MG-17, NARMA-10, and MSTSN, respectively. These results indicate that the PSO-based network topology optimization method is more effective than the GA-based network topology for an MSCR system on time-series prediction tasks.
\par
Finally, the network topologies corresponding to the best-performing MSCR-PSO models are visualized in Figure~\ref{fig:searched_topologies}. Note that all directed connections originating from invalid vertex-encoders are omitted, as these vertex-encoders produce only zero state vectors, regardless of their connections to valid vertex-encoders. After optimization with PSO, the number of valid vertex-encoders in MSCR-PSO is reduced to seven, five, and one for the MG-17, NARMA-10, and MSTSN datasets, respectively. 
Notably, we observed that the optimal solution for MSTSN is a rank-one MSCR, which by Remark~\ref{rmk:rank} effectively corresponds to a single SCR. This structural change highlights the flexibility of MSCR in adapting its network topology to specific problems, allowing it to simplify into a single SCR when necessary.
\begin{table}[htbp!]
\centering
\caption{The Best average RMSEs with the corresponding standard deviations for all the tested models on the MG-17 dataset}
\label{tab: MG-17}
\begin{tabular}{cccc}
\hline
Model      & Distribution & $f\left ( \cdot  \right ) $ & RMSE$\pm$(STD)          \\ \hline
SCR        & Bernoulli    & Tanh                & 2.40E-02$\pm$(2.26E-03) \\
DeepSCR    & Bernoulli    & Identity            & 4.44E-04$\pm$(7.20E-06) \\
GroupedSCR & Bernoulli    & Tanh                & 2.22E-02$\pm$(2.21E-03) \\
MSCR-GA    & Bernoulli    & Identity            & 1.69E-03$\pm$(6.10E-04) \\
MSCR-PSO & Bernoulli & Tanh & \textbf{2.50E-04}$\pm$(\textbf{7.37E-05}) \\ \hline
\end{tabular}
\end{table}

\begin{table}[htbp!]
\centering
\caption{The Best average RMSEs with the corresponding standard deviations for all the tested models on the NARMA-10 dataset}
\label{tab: NARMA-10}
\begin{tabular}{cccc}
\hline
Model   & Distribution & $f\left ( \cdot  \right ) $    & RMSE$\pm$(STD)          \\ \hline
SCR    & Bernoulli    & Tanh     & 3.87E-02$\pm$(5.95E-05) \\
DeepSCR  & Bernoulli    & Tanh   & 3.90E-02$\pm$(4.81E-04)  \\
GroupedSCR & Bernoulli   & Tanh & 2.31E-02$\pm$(1.61E-03) \\
MSCR-GA  & Bernoulli    & Identity   & 4.09E-02$\pm$(1.67E-04) \\
MSCR-PSO & Bernoulli    & Tanh   & \textbf{1.69E-02}$\pm$(\textbf{4.31E-03}) \\ \hline
\end{tabular}
\end{table}

\begin{table}[htbp!]
\centering
\caption{The Best average RMSEs with the corresponding standard deviations for all the tested models on the MSTSN dataset}
\label{tab: MSTSN}
\begin{tabular}{cccc}
\hline
Model & Distribution & $f\left ( \cdot  \right ) $   & RMSE$\pm$(STD)       \\ \hline
SCR   & Bernoulli    & Identity  & 4.17E-03$\pm$(4.75E-06)        \\
DeepSCR  & Bernoulli    & Identity & 4.16E-03$\pm$(1.35E-05)  \\
GroupedSCR & Bernoulli    & Identity & \textbf{4.11E-03}$\pm$(\textbf{3.24E-06}) \\
MSCR-GA & Bernoulli    & Identity & 4.19E-03$\pm$(8.17E-06) \\
MSCR-PSO & Bernoulli    & Identity &  4.12E-03$\pm$(1.13E-06) \\ \hline
\end{tabular}
\end{table}

\section{Discussion}
\label{sec:discussion}

In this work, we introduced the notion of MSCR by designing a multi-reservoir system composed of small SCR vertex-encoders. We optimized both the input scaling factors and the network topology governing the interconnectivity between SCR vertex-encoders, resulting in the MSCR-PSO model. MSCR-PSO was compared against SCR, MSCR with hand-crafted network topologies, and MSCR optimized using an existing method for Multi-RC~\cite{li2024designing}.

Across the three benchmark datasets, we found that MSCR-PSO achieves competitive predictive performance while utilizing a lower-dimensional state space compared to both SCR and MSCR with pre-determined network topologies. The significance of this work lies in the advancements of MSCR-PSO over hand-crafted MSCR, which paves the way for systemic optimization of task-specific multi-reservoir systems.

Furthermore, since PSO operates on continuous values, it can achieve both of the following simultaneously:
\begin{itemize}
    \item direct optimization of tune real‑valued entries, such as $\H$ and $\s$, and
    \item optimize the binary entries of both $\A$ and $\d$ via specific transformation functions.
\end{itemize}
In contrast, GA relies on discrete encoding and is inherently limited to categorical or binary search. This flexibility of operating on continuous values makes PSO particularly well‑suited for constructing MSCR. Experimental results also empirically demonstrate its superiority over existing GA-based optimization methods for optimizing multi-reservoir systems.

In this work, we achieved competitive performance with MSCR solely by optimizing network topologies and input scaling factors. In future works, we aim to further explore the computational potential of MSCR by extending the optimization process to include additional hyperparameters, such as spectral radius and regularization parameter.

\section*{Acknowledgment}
This work was partly supported by JSPS KAKENHI Grant Number JP23K28154 (GT), JST CREST Grant Numbers JPMJCR19K2 (ZL, KF, GT), JPMJCR24R2 (GT), and JST Moonshot R\&D Grant Number JPMJMS2021 (ZL, KF, KA, GT), Cross-ministerial Strategic Innovation Promotion Program (SIP), the 3rd period of SIP, Grant Numbers JPJ012207 (KA), JPJ012425 (KA, KF),
Institute of AI and Beyond of UTokyo (KA), the World Premier International Research Center Initiative (WPI), MEXT, Japan.



\end{document}